\documentclass[letterpaper, 10 pt, conference]{ieeeconf}  

\IEEEoverridecommandlockouts      

\usepackage{times}
\usepackage{booktabs}
\usepackage{balance}
\usepackage{siunitx}
\usepackage{multicol}
\usepackage[utf8]{inputenc}
\usepackage{graphicx}
\usepackage{lipsum}
\usepackage{wrapfig}
\usepackage[ruled,vlined,linesnumbered,boxed]{algorithm2e}
\usepackage[colorlinks=true, urlcolor=blue, linkcolor=black]{hyperref} 
\usepackage{cleveref}
\usepackage{soul}
\usepackage{todonotes}
\usepackage{amsmath, amssymb}
\usepackage{bm}
\usepackage{color}
\usepackage{multirow}

\usepackage[caption=false,font=footnotesize]{subfig}
\usepackage{bbm}

\title{\LARGE \bf
Dynamic Input for Deep Reinforcement Learning \\in Autonomous Driving
}

\author{Maria Huegle$^{1,*}$, Gabriel Kalweit$^{1,*}$, Branka Mirchevska$^{2}$, Moritz Werling$^{2}$, Joschka Boedecker$^{1,3}$\\
\thanks{$^{*}$Authors contributed equally.}%
\thanks{$^{1}$Dept. of Computer Science, University of Freiburg, Germany.}%
\thanks{{\tt \{hueglem,kalweitg,jboedeck\}@cs.uni-freiburg.de}}
\thanks{$^{2}$BMWGroup, Unterschleissheim, Germany.}%
\thanks{{\tt  \{Branka.Mirchevska,Moritz.Werling\}@bmw.de}}%
\thanks{$^{3}$Cluster of Excellence BrainLinks-BrainTools, Freiburg, Germany.}%
}

\newcommand\copyrighttext{%
  \footnotesize \textcopyright 2019 IEEE. Personal use of this material is permitted. Permission from IEEE must be obtained for all other uses, in any current or future media, including reprinting/republishing this material for advertising or promotional purposes, creating new collective works, for resale or redistribution to servers or lists, or reuse of any copyrighted component of this work in other works.}

\newcommand\copyrightnotice{%
\begin{tikzpicture}[remember picture,overlay]
\node[anchor=south,yshift=15pt] at (current page.south) {\fbox{\parbox{\dimexpr\textwidth-\fboxsep-\fboxrule\relax}{\copyrighttext}}};
\end{tikzpicture}%
}

\begin{document}

\maketitle
\copyrightnotice

\thispagestyle{empty}
\pagestyle{empty}

\begin{abstract}
In many real-world decision making problems, reaching an optimal decision requires taking into account a variable number of objects around the agent. Autonomous driving is a domain in which this is especially relevant, since the number of cars surrounding the agent varies considerably over time and affects the optimal action to be taken. Classical methods that process object lists can deal with this requirement. However, to take advantage of recent high-performing methods based on deep reinforcement learning in modular pipelines, special architectures are necessary. For these, a number of options exist, but a thorough comparison of the different possibilities is missing. In this paper, we elaborate limitations of fully-connected neural networks and other established approaches like convolutional and recurrent neural networks in the context of reinforcement learning problems that have to deal with variable sized inputs. We employ the structure of Deep Sets in off-policy reinforcement learning for high-level decision making, highlight their capabilities to alleviate these limitations, and show that Deep Sets not only yield the best overall performance but also offer better generalization to unseen situations than the other approaches.
\end{abstract}

\section{Introduction}
Many autonomous driving systems are built upon a modular pipeline consisting of perception, localization, mapping, high-level decision making and motion planning. The perception component extracts a list of surrounding objects and traffic participants like vehicles, pedestrians and bicycles. The number of objects that are relevant for the later decision making process can be highly dynamic, in both highway and urban scenarios. Classical rule-based decision-making systems can process these lists directly, but are limited due to their fragility in light of ambiguous and noisy sensor data. Deep Reinforcement Learning (DRL) methods offer an attractive alternative for learning decision policies from data automatically and have shown great potential in a number of domains \cite{DBLP:journals/nature/MnihKSRVBGRFOPB15, DBLP:journals/nature/SilverHMGSDSAPL16, DBLP:conf/nips/WatterSBR15, DBLP:journals/jmlr/LevineFDA16}. Promising results were also shown for learning driving policies from raw sensor data~\cite{DBLP:conf/icra/JaritzCTPN18}. However, end-to-end methods can suffer from a lack of interpretability and can be difficult to train. These issues can be alleviated by learning from low-dimensional state features extracted by a perception module.
To take advantage of DRL methods in scenarios with dynamic input lengths, though, special architectural components are necessary as part of the learning system. Several neural network based options allow processing of variable-sized input, as detailed below. However, it is not clear which of these is superior in context of a DRL algorithm, enabling high performance and generalization to various scenarios. 

\begin{figure}[t]
    \centering
    \vspace*{0.3cm}
    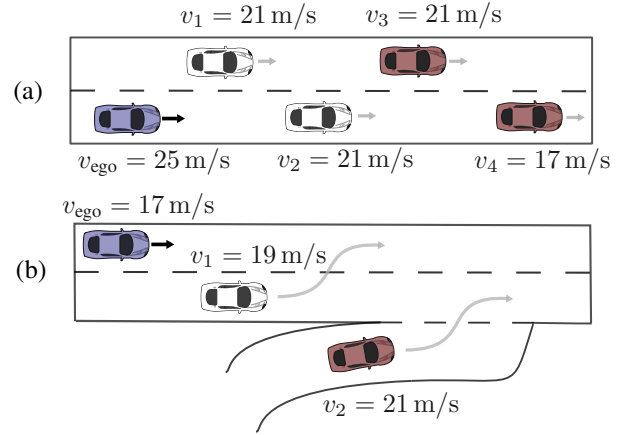
    \caption{Limitations of fixed input representations consisting of (a) the $2$-nearest surrounding vehicles, where vehicles 3 and 4 are not visible  (b) a relational grid with $\Delta_{\text{lateral}} = 1$ considered side lanes. Cars considered by the input representation are shown in white, invisible cars in red and the agent in blue.}
    \label{fig:motivation_deepsets}
\end{figure}

Prior research mostly relied on fixed sized inputs \cite{adaptive_behavior_kit, branka_rl_highway, 2018TowardsPH, overtaking_maneuvers_kaushik} or occupancy grid representations \cite{tactical_decision_making, deeptraffic} for value or policy estimation. However, fixed sized inputs limit the number of considered vehicles, for example to the $n$-nearest vehicles. \Cref{fig:motivation_deepsets} (a) shows a scenario where this kind of representation is not enough to make the optimal lane change decision, since the agent does not represent the $n+1$ and $n+2$ closest cars even though they are within sensor range. A more advanced fixed sized representation  \cite{adaptive_behavior_kit, branka_rl_highway} considers a relational grid of $\Delta_{\text{ahead}}$ leaders and $\Delta_{\text{behind}}$ followers in $\Delta_{\text{lateral}}$ side lanes around the agent. As \Cref{fig:motivation_deepsets} (b) shows, this can still be insufficient for optimal decision making, since the white car on the incoming $\Delta_{\text{lateral}}+1$ right lane is not within the maximum of considered lanes, and therefore has no influence on the decision making process.

Using occupancy grids in combination with convolutional neural networks (CNN) imposes a trade-off between computational workload and expressiveness. Whilst smaller architectures acting on low-resolution grids as an input are efficient from a computational perspective, they may be too imprecise to represent the environment correctly. Whereas for high-resolution grids, most computations can be potentially redundant due to the sparsity of grid maps. In addition, a grid is limited to its initially defined size. Objects off the grid cannot be represented and are therefore not considered in decision making.

A third way to deal with variable input sequences are recurrent neural networks (RNNs). Even though the temporal context might be important for predictions w.r.t. a single or a group of objects, the order in which the objects are fed into the value function or policy estimator should not be relevant for a fixed time step -- i.e., the input should be \textit{permutation invariant}. Combined with an attention mechanism, a recurrent network can be used to create a set representation which is permutation invariant w.r.t. the input elements \cite{2015arXiv151106391V}, which we subsequently call Set2Set. However, RNNs tend to be difficult to train \cite{Pascanu:2013:DTR:3042817.3043083}, e.g. due to vanishing and exploding gradients, which can be a problem in highly stochastic environments.

Lastly, Deep Sets can be employed to process inputs of varying size \cite{NIPS2017_6931}. The Deep Set architecture was used for point cloud classification and anomaly detection \cite{NIPS2017_6931}, as well as in a sim-to-real DQN-setting for a robot sorting task \cite{DBLP:journals/corr/abs-1809-07480}.

In this paper, we suggest to use Deep Sets for deep reinforcement learning as a flexible and permutation invariant architecture to handle a variable number of surrounding vehicles that influence the decision making process for lane-change maneuvers. 
We additionally propose to use the Set2Set architecture as both baseline and another new approach to deal with a variable input size in reinforcement learning. Our main contributions are the formalization of the DeepSet-Q and Set2Set-Q algorithms, and the extensive evaluation of DeepSet-Q, comparing it to various other approaches to handle dynamic inputs, and showing that Deep Sets outperform related approaches in the application of high-level decision making in autonomous lane changes, while generalizing better to unseen situations.

\section{Method}
In reinforcement learning, an agent acts in an environment by applying action $a_t \sim \pi$, following policy $\pi$, in \mbox{state $s_t$}, gets into some state $s_{t+1}\sim\mathcal{M}$, according to model $\mathcal{M}$, and receives a reward $r_{t}$ in each discrete time step $t$. The agent has to adjust its policy $\pi$ to maximize the discounted long-term return $R(s_t) = \sum_{t'>=t} \gamma^{t'-t}r_{t'}$, where $\gamma \in [0, 1]$ is the discount factor. 
We focus on the case, where we find the optimal policy based on model-free Q-learning \cite{Watkins92q-learning}. The Q-function $Q^\pi(s_t,a_t)=\mathbf{E}_{a_{t'>t}\sim\pi}[R(s_t)|a_t]$ represents the value of an action $a_t$ and following $\pi$ thereafter. From the optimal action-value function $Q^*$, we can then easily extract the optimal policy by maximization.\\

\subsection{DeepSet-Q}
 In the DeepSet-Q approach, 
 we train a neural network $Q_{\mathcal{DS}}(\cdot, \cdot|\theta^{Q_{\mathcal{DS}}})$, parameterized by $\theta^{Q_{\mathcal{DS}}}$, to estimate the action-value function via DQN \cite{DBLP:journals/nature/MnihKSRVBGRFOPB15} for state \mbox{$s_t=(X^{\text{dyn}}_{t}, x_{t}^{\text{static}})$} and action $a_t$. The state consists of a dynamic input \mbox{$X_t^{\text{dyn}}=[x^1_t, .., x^{\text{seq len}}_t]^\top$} with a variable number\footnote{In the application of autonomous driving, the sequence length is equal to the number of vehicles in sensor range.} of vectors $x^j_t|_{0\le j \le \text{seq len}}$ and a static input vector $x_{t}^{\text{static}}$.
 
\begin{figure}
    \center
    \includegraphics[width=8cm]{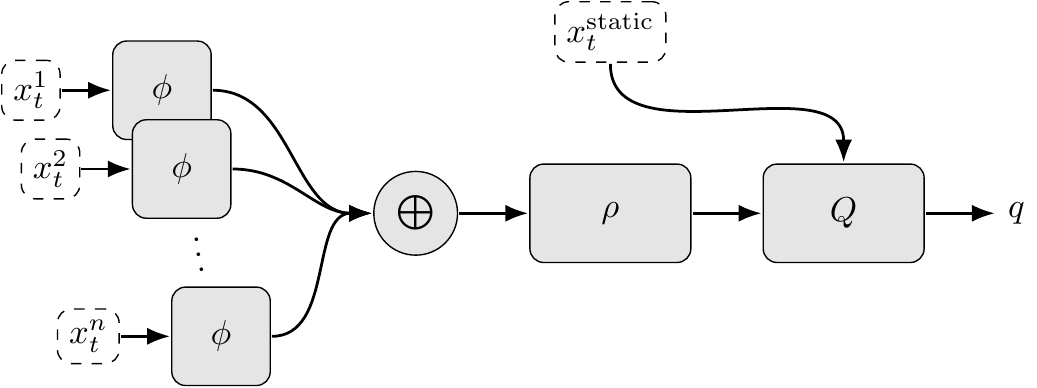}
    \caption{Scheme of DeepSet-Q. The modules $\phi$, $\rho$ and $Q$ are fully-connected neural networks and  $\bigoplus$ is a permutation invariant pooling operation. Vectors $x^j_t$ and $x^{\text{static}}_t$ are dynamic and static feature representations at time step $t$, $q$ is the action-value output.}
    \label{fig:dsscheme}
\end{figure}

 The Q-network $Q_{\mathcal{DS}}$ consists of three main parts, $(\phi,\rho,Q)$. The input layers are built of two neural networks $\phi$ and $\rho$, which are components of the Deep Sets. The representation of the input set is computed by: $$\Psi(X_t^{\text{dyn}}) = \rho\left(\sum_{x^j_t\in X_t^{\text{dyn}}}\phi(x_t^j)\right),$$ which makes the Q-function permutation invariant w.r.t. its input \cite{NIPS2017_6931}. An overview of the Q-function is shown in \Cref{fig:dsscheme}. Instead of taking the \textit{sum}, other permutation invariant pooling functions, such as the \textit{max}, can be used. Static feature representations $x_t^{\text{static}}$ can be fed directly to the \mbox{$Q$-module}. The final Q-values are then given by $Q_{\mathcal{DS}}(s_t, a_t)=Q([\Psi(X_t^{\text{dyn}}), x_{t}^{\text{static}}], a_t)$, where $[\cdot, \cdot]$ denotes concatenation.
The Q-function is trained on minibatches sampled from a replay buffer $\mathcal{R}$, which contains transitions collected by some policy $\hat{\pi}$. We then minimize the loss function: 
$$L(\theta^{Q_{\mathcal{DS}}}) = \frac{1}{b}\sum\limits_i\left(y_i - Q_{\mathcal{DS}}(s_i, a_i)\right)^2,$$ with targets $y_i=r_i + \gamma \max_{a} Q'_{\mathcal{DS}}(s_{i+1}, a),$
where $(s_i, a_i, s_{i+1}, r_i)|_{0\leq i \leq b}$ is a randomly sampled minibatch from the replay buffer. The target network is a slowly updated copy of the Q-network.
In every time step, the parameters of the target network $\theta^{Q'_{\mathcal{DS}}}$ are moved towards the parameters of the Q-network by step-length $\tau$, i.e. $\theta^{Q'_{\mathcal{DS}}}\leftarrow\tau\theta^{Q_{\mathcal{DS}}} + (1-\tau)\theta^{Q'_{\mathcal{DS}}}$. For a detailed description, see \Cref{alg:setq}. To overcome the problem of overestimation, we further apply a variant of Double-Q-learning which is based on two independent network pairs, so as to use the minimum of the predictions for target calculation \cite{DBLP:conf/icml/FujimotoHM18}.

\begin{algorithm}[t]
    \SetAlgoLined
    \DontPrintSemicolon
    initialize $Q_{\mathcal{DS}}=(\phi,\rho,Q)$ and $Q'_{\mathcal{DS}}=(\phi',\rho',Q')$\\
    set replay buffer $\mathcal{R}$\\

    \For{\text{optimization step} o=1,2,\dots}{
        get minibatch $(s_i,a_i,(X^{\text{dyn}}_{i+1}, x_{i+1}^{\text{static}}),r_{i})$ from $\mathcal{R}$\\
        \ForEach{\text{transition}}{
            \ForEach{\text{object }$x_{i+1}^j$ in $X^{\text{dyn}}_{i+1}$}{
                $(\phi'_{i+1})^j=\phi'\left(x_{i+1}^j\right)$\\
            }
            $\rho'_{i+1}=\rho'\left(\sum\limits_j(\phi'_{i+1})^j\right)$\\
             
            $y_i = r_i+\gamma\max_a Q'(\rho'_{i+1}, x_{i+1}^{\text{static}}, a)$\\
        }
        perform a gradient step on loss: $$\frac{1}{b}\sum\limits_i\left(Q_{\mathcal{DS}}(s_i, a_i)-y_i\right)^2$$\\
        update target network by:
        $$\theta^{Q_{\mathcal{DS}}'}\leftarrow\tau\theta^{Q_{\mathcal{DS}}} + (1-\tau)\theta^{Q_{\mathcal{DS}}'}$$
    }
    \vspace{0.2475cm}
    \caption{DeepSet-Q}
    \label{alg:setq}
\end{algorithm}

\subsection{Set2Set-Q}

\begin{figure}
    \center
    \includegraphics[width=8cm]{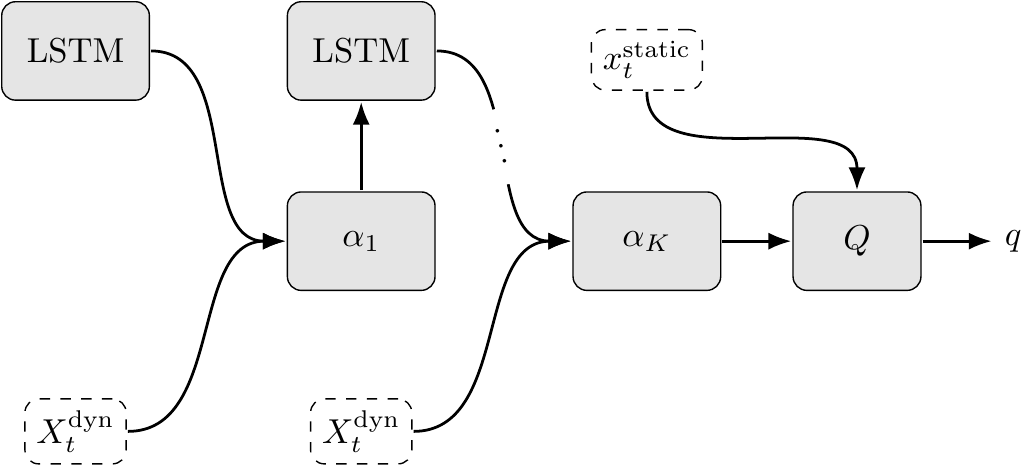}
    \caption{Unrolled scheme of Set2Set-Q. Dynamic feature representations $X_t^{\text{dyn}}=[x^1_t, .., x^{\text{seq len}}_t]^\top$ are fed via a permutation invariant attention mechanism $\alpha_k$ to the LSTM in $K$ iterations and accumulated in $q_k^*$. Vector $x^{\text{static}}_t$  is the static feature representation at time step $t$. $q$ is the action-value output.}
    \label{fig:s2sscheme}
\end{figure}

Another approach to deal with a variable number of input elements is to replace $Q_{\mathcal{DS}}$ and target network $Q'_{\mathcal{DS}}$ by recurrent network architectures $Q_{S2S}$ and $Q'_{S2S}$. Here, the input layer of the Q-network consists of the recurrent module proposed in  \cite{2015arXiv151106391V}, which uses an aggregation in a hidden state of a long-term short-term memory (LSTM) \cite{Hochreiter:1997:LSM:1246443.1246450} as a pooling operation for the input elements. Combined with their proposed attention mechanism, the aggregation of the input set is permutation invariant. The Q-network consists of two main parts $(\text{LSTM}, Q)$. An overview is shown in \Cref{fig:s2sscheme}. The LSTM is updated $K$ times, where $k$ denotes the iteration index. It evolves a state $q_k^*$ with  $q_k = \text{LSTM}(q^*_{k-1})$, which is formed by a concatenation $ q_k^* = [q_k, \beta_k]$ of a query vector $q_k$ with the readout vector $ \beta_k =  \sum_j \alpha_{j, k}  x^j$. The readout vector is computed over all input set elements $x^j \in X^{\text{dyn}}$ multiplied by an attention factor $ \alpha_{j,k}(x^j, q_k)$. The resulting state representation of the dynamic objects $q^*_K$ is concatenated with static feature representation $x^{\text{static}}$ and fed into the $Q$-module.  For details, see \Cref{alg:set2setq}.

\begin{algorithm}
    \SetAlgoLined
    \DontPrintSemicolon
      initialize $Q_{S2S}=(\text{LSTM}, Q)$ and $Q'_{S2S}=(\text{LSTM}', Q')$\\
    set replay buffer $\mathcal{R}$\\
        \For{\text{optimization step} o=1,2,\dots}{
            get minibatch $(s_i,a_i,(X^{\text{dyn}}_{i+1}, x_{i+1}^{\text{static}}),r_{i})$ from $\mathcal{R}$\\
            \ForEach{\text{transition }}{
                reset $hidden$ state of LSTM$'$ and $q_0^*$\\
                 \For{\text{iteration} k=1,2,\dots, K}{
                       $q_k = \text{LSTM}'(q^*_{k-1})$\\
                        \ForEach{\text{object }$x_{i+1}^j$ in $X^{\text{dyn}}_{i+1}$}{ 
                              $e_{j,k} = x_{i+1}^j \cdot q_k$\\
                       }
                     $\displaystyle \alpha_{j,k} = \frac{exp(e_{j,k})}{\sum_m exp(e_{m,k})}$\\
                       $\displaystyle \beta_k =  \sum_j \alpha_{j, k}  x_{i+1}^j$\\
                    $ q_k^* = [q_k, \beta_k]$
                    }
                 $y_i = r_{i}+\gamma\max_a Q'(q_K^*, x_{i+1}^{\text{static}}, a)$\\
             }
        perform gradient step and update target network as in \Cref{alg:setq}
        }
    \caption{Set2Set-Q}
    \label{alg:set2setq}
\end{algorithm}

\section{Experimental Setup}
We apply the Deep Set and Set2Set input representations in the reinforcement learning setting for high-level decision making in autonomous driving.

\subsection{Application to Autonomous Driving}
In order to model this task as a MDP, we first define state space, action space and reward function.

\subsubsection{State Space} \label{state_space}

For the agent itself, subsequently called ego vehicle, we use the absolute velocity $v_{\text{ego}}\in\mathbb{R}_{\geq0}$ and whether lanes to the left and right of the agent are available or not. For all vehicles $j$ within the scope of an input representation and within the maximum sensor range $d_\text{max}$, we consider the following features: 
\begin{itemize}
\item the relative distance $dr_j = (p_j - p_{\text{ego}})/d_\text{max}\in\mathbb{R}$, where $p_{\text{ego}}$,  $p_j$ are longitudinal positions in a curvilinear coordinate system of the lane,
\item the relative velocity $dv_j = \frac{v_j-v_{\text{ego}}}{v_{\text{ego}} + \epsilon} \in\mathbb{R}$, where $\epsilon \ll 1$ and  $v_{\text{ego}}$, $v_j$ denote absolute velocities, 
\item and the relative lane\footnote{e.g. $dl_c=-1$ for vehicle $c$ which is one lane to the left of the agent. This feature is not needed for the fixed input and occupancy grids.} $dl_j = l_j - l_{\text{ego}} \in \mathbb{N}$ where $l_j$, $l_{\text{ego}} $ are lane indices.
\end{itemize}

\subsubsection{Action Space}

The action space consists of a discrete set of possible actions $\mathcal{A} = \{$keep lane, perform left lane-change, perform right lane-change$\}.$ The actions are high-level decisions of the agent in lateral direction. Acceleration towards reaching the desired velocity is controlled by a low-level execution layer. Collision avoidance and maintaining safe distance to the preceding vehicle are handled by an integrated safety module, analogously to \cite{deeptraffic}, \cite{Mirchevska2018HighlevelDM}. In our case the agent simply keeps the lane, if the chosen action is not safe.\\

\subsubsection{Reward Function}
For a desired velocity of the agent $v_\text{desired}$, we define the reward function\footnote{In favor of simplicity, we omit considering more factors such as jerk or cooperativeness.} $r: \mathcal{S}\times\mathcal{A}\mapsto \mathbb{R}$ as:
$$r(s, a) = 1 - \frac{|v_{\text{ego}}- v_{\text{desired}}|}{v_{\text{desired}}} - p_{\text{lc}}, $$
where $p_{\text{lc}}$ is a penalty for choosing a lane change action. In our experiments, we use $p_{\text{lc}} = 0.01$ if action $a$ is a lane change and $0$ otherwise. The weight was chosen empirically in preliminary experiments.

\subsection{Input Representations}
In this section, we describe how to transform the given state information to three different input representations. In total, we trained and evaluated four DQN agents with different input modules. For all approaches, including the baselines, we optimized the hyperparameters extensively. A detailed overview of the configuration spaces can be found in the appendix.\\

\subsubsection{Relational Grid (Fixed Input)} As fixed input representation, we use a relational grid with neighborhood $\Delta_{\text{ahead}}=\Delta_{\text{behind}}=\Delta_{\text{lateral}}=2$, which results in a maximum of $20$ surrounding cars\footnote{In preliminary experiments we further investigated a relational grid with $\Delta_{\text{ahead}}=\Delta_{\text{behind}}=\Delta_{\text{lateral}}=3$, resulting in 42 considered vehicles. However, the results showed very high variance. Additionally, we omitted experiments with 6 vehicles because of the limitations shown in \Cref{fig:motivation_deepsets}.}. For non-existent cars in the grid we use default value $dv = 0$, as well as $dr=1$ for leaders and $dr = -1$ for followers. These values correspond to a vehicle with same speed at maximum sensor range, having no influence on the decisions of the agent. In total, the state consists of $43$ input features.\\

\subsubsection{Occupancy Grid}
A two-dimensional occupancy grid depicts the scene around the ego vehicle from a bird's eye perspective as an input. We use a grid size\footnote{We also evaluated a grid size of $160 \times 5$. In our experiments, however, the larger grid led to a worse performance after the same number of gradient steps.} of $80 \times 5$, where the ego vehicle is represented in the middle. We therefore observe $5$ highway lanes. All cells in the grid occupied by a surrounding vehicle $j$ are assigned values $1 + dv_j$ and the cells occupied by the ego vehicle $dv_{\text{ego}}=1$. Free cells are assigned zeros.\\

\subsubsection{Set Input}
The input for DeepSet-Q and Set2Set-Q consists of a variable length list of feature vectors. If no vehicles are in sensor range, the forward pass of $\phi$ is omitted for the Deep Sets and a vector of zeros is fed to $\rho$. In the Set2Set model, the forward passes of the LSTM are skipped and $q_K^*$ is set to zero.

\subsection{SUMO Simulator}

\begin{figure}[t]
    \centering
    \includegraphics[width=0.5\textwidth]{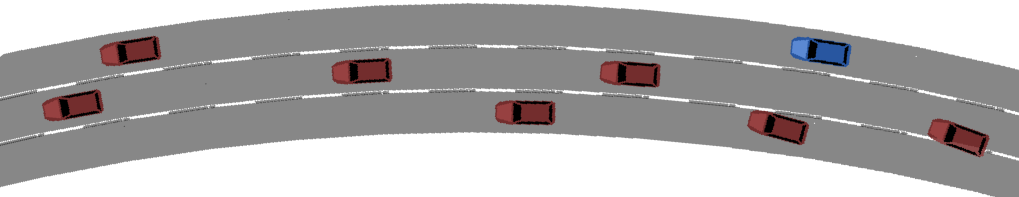}
    \caption{Birdview of the 3-lane highway scenario in SUMO. The ego vehicle is shown in blue. All other vehicles are shown in red. }
    \label{fig:sumo_ablation}
\end{figure}

For experiments, we used the open-source SUMO traffic simulation framework \cite{SUMO}, as shown in \Cref{fig:sumo_ablation}. Training and evaluation scenarios were performed on a simulated $1000\,$m circular highway with three lanes. The vehicles were randomly positioned passenger cars with different driving behaviors to resemble realistic highway traffic as much as possible. We set the sensor range to $\SI{80}{\metre}$ in front and behind the ego vehicle. Details for the simulation environment and driving behaviors are shown in appendix, \Cref{tab:driver_types}. 

\subsection{Training Setup}
\label{subseq:setup}
The data was collected in SUMO simulation scenarios. We trained our agents offline on two separate datasets:

\begin{itemize}
\item  Dataset 1:
500.000 transition samples with an arbitrary number  of  surrounding  vehicles  in  each  transition. 
\item Dataset 2:
500.000  transition  samples with  at most  six  surrounding  vehicles.  This dataset was sampled  proportional to the original data distribution.
\end{itemize}

The datasets were collected by a data-collection agent in traffic scenarios with a random number of  $n \in [30, 60]$ vehicles. The  data-collection agent had the SUMO safety module integrated and
performed a random lane change to the left or right whenever possible. 

\begin{figure}[t]
    \centering
    \includegraphics[width=0.5\textwidth]{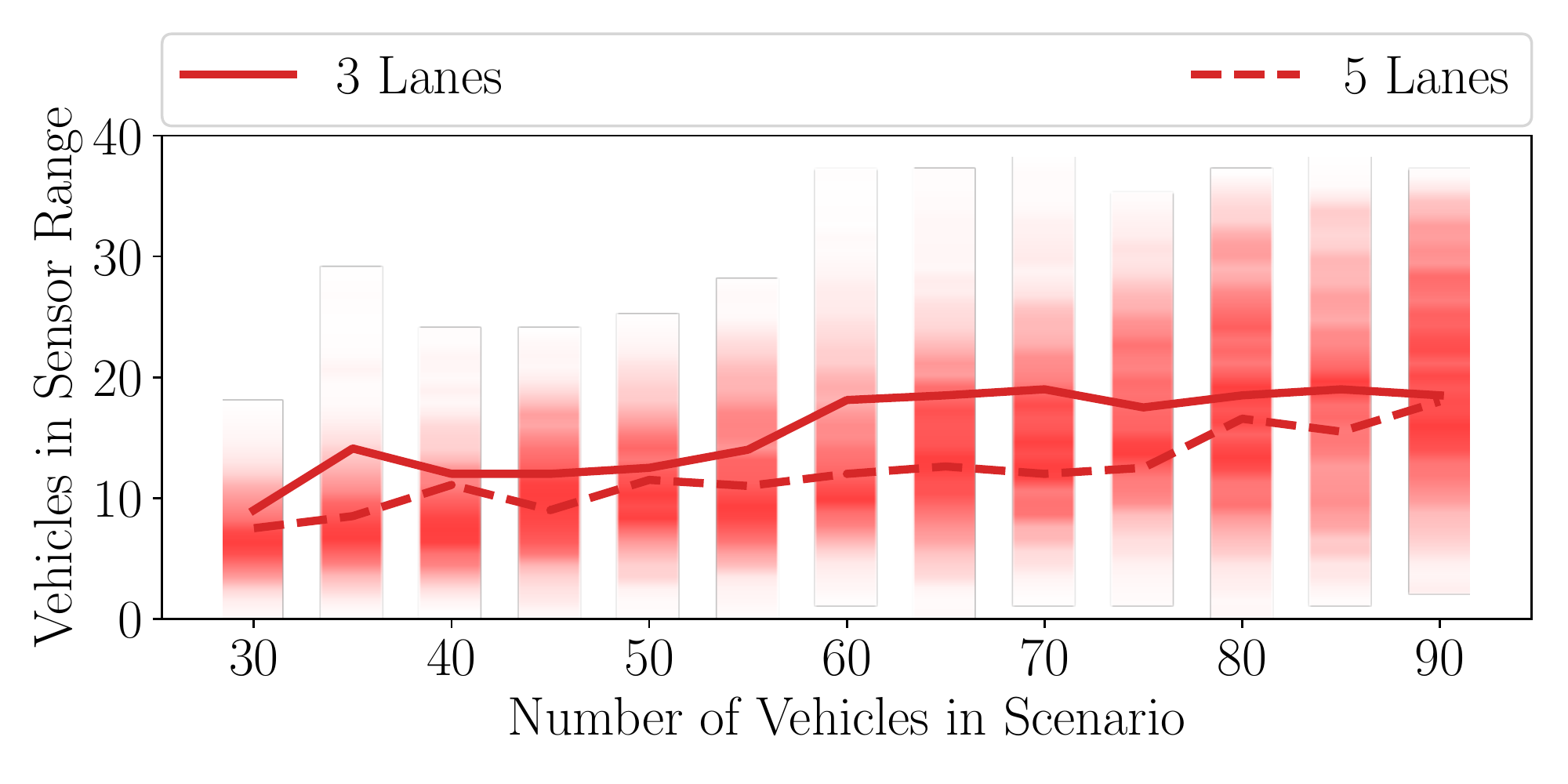}
    \caption{The vertical bars denote the distribution of transitions for a given number of vehicles in sensor range, shown for scenarios with 30 to 90 vehicles on three lanes. The darker, the higher the frequency. Lines represent the mean number of vehicles in sensor range for three lanes (solid) and five lanes (dashed).}
    \label{fig:surr_vehicles}
\end{figure}

\begin{figure*}[t!]
    \centering
    \subfloat[\label{fig:main}]{%
     \includegraphics[width=0.4\textwidth,trim=0cm 0cm 0cm 2cm, clip]{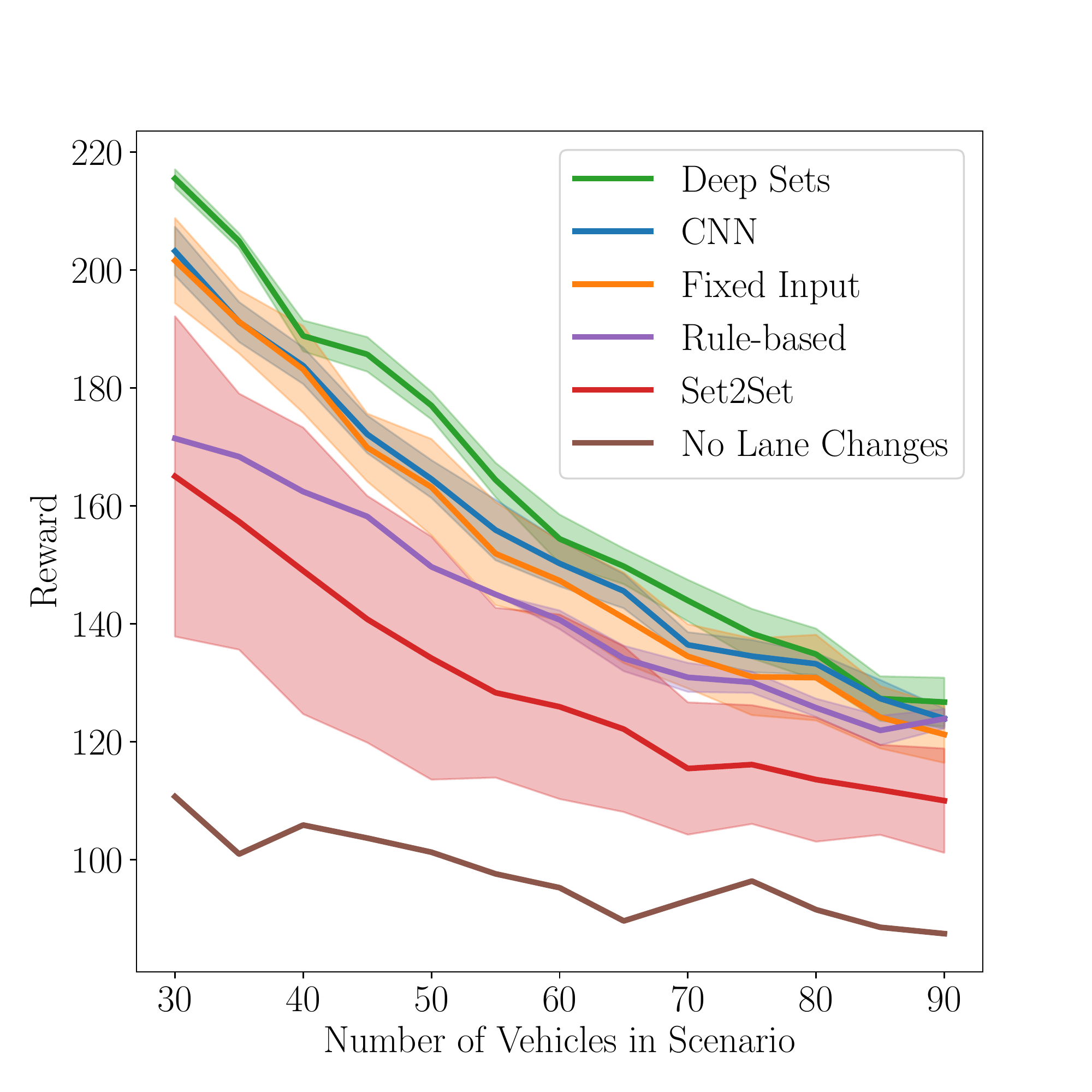}
     \hfill
    }
    \subfloat[\label{fig:gen}]{%
     \includegraphics[width=0.4\textwidth,trim=0cm 0cm 0cm 2cm, clip]{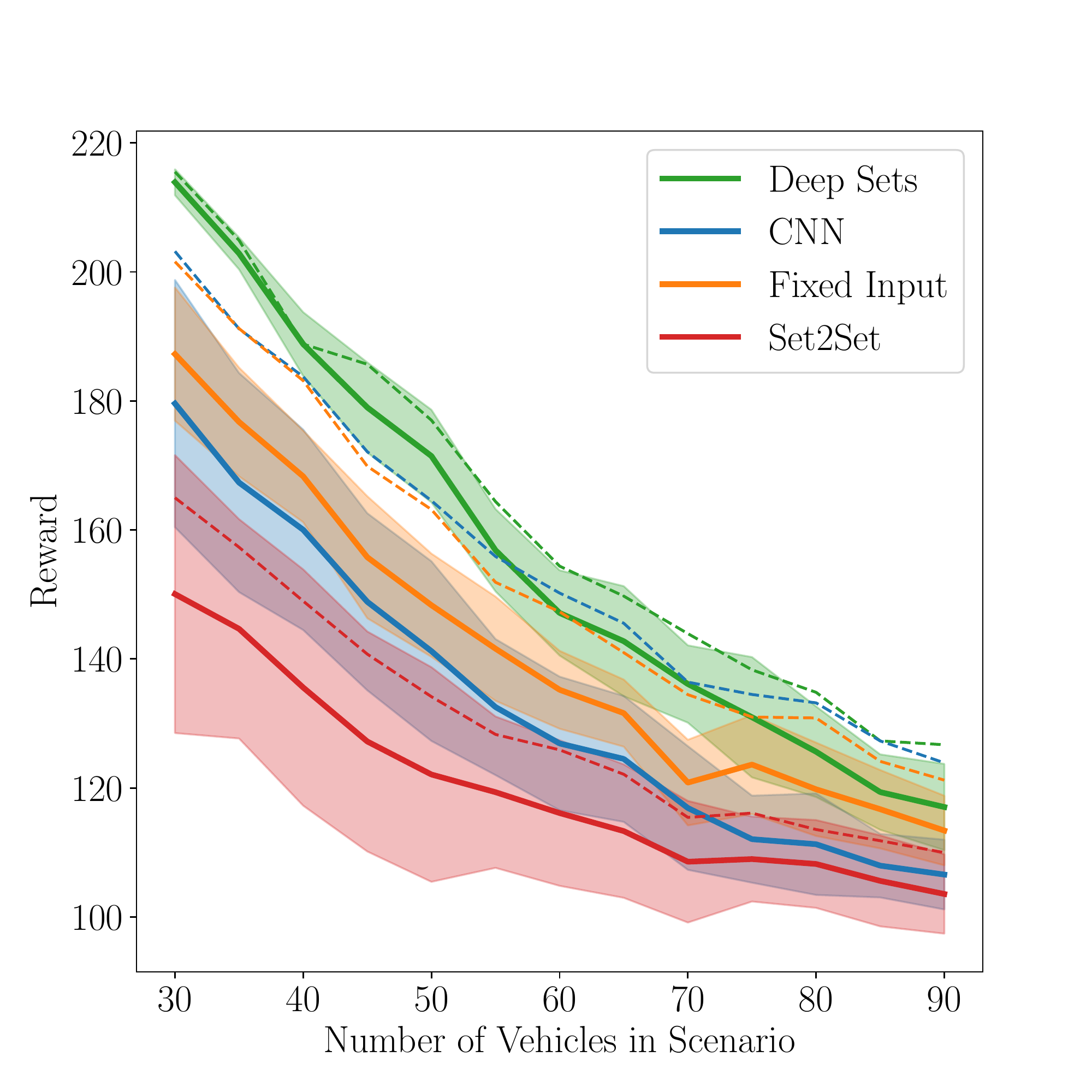}
     \hfill
    }
\caption{(a) Mean performance and standard deviation over 10 training runs on Dataset 1. (b) Mean performance and standard deviation over 10 training runs, trained on Dataset 2 (solid) and on Dataset 1 (dashed). 
}
\end{figure*}

\begin{table*}[t!]
    \centering
    \begin{tabular}{ccccc}
    	\toprule
    	Scenario&30&50&70&90\\
    	\midrule
    	Fixed Input & 201.59 ($\pm$ 7.23) & 163.17 ($\pm$ 8.17) & 134.47 ($\pm$ 5.45) & 121.21 ($\pm$ 4.83)\\
    	CNN & 203.2 ($\pm$ 4.15) & 164.54 ($\pm$ 3.2) & 136.41 ($\pm$ 2.15) & 123.9 ($\pm$ 1.7)\\
    	Set2Set & 164.99 ($\pm$ 27.17) & 134.15 ($\pm$ 20.61) & 115.43 ($\pm$ 11.22) & 109.98 ($\pm$ 8.85)\\
    	\midrule
    	Deep Sets & \textbf{215.51} ($\pm$ 1.58) & \textbf{177.01} ($\pm$ 2.35) & \textbf{143.93} ($\pm$ 3.52) & \textbf{126.7} ($\pm$ 4.14)\\
    	\bottomrule
    \end{tabular}

    \caption{Mean performance and standard deviation of the approaches trained on samples with all vehicles (Dataset 1).}
    \label{tab:res}
\end{table*}
\begin{table*}
    \centering
    \begin{tabular}{ccccc}
    	\toprule
    	Scenario&30&50&70&90\\
    	\midrule
    	Fixed Input & 187.26 ($\pm$ 10.33) & 148.33 ($\pm$ 8.0) & 120.83 ($\pm$ 6.64) & 113.4 ($\pm$ 5.4)\\
    	Set2Set & 150.06 ($\pm$ 21.55) & 122.09 ($\pm$ 16.64) & 108.56 ($\pm$ 9.43) & 103.56 ($\pm$ 6.15)\\
    	CNN & 179.6 ($\pm$ 19.18) & 141.21 ($\pm$ 13.93) & 116.89 ($\pm$ 9.59) & 106.57 ($\pm$ 5.41)\\
    	\midrule
    	Deep Sets & \textbf{213.91} ($\pm$ 2.02) & \textbf{171.46} ($\pm$ 7.17) & \textbf{136.11} ($\pm$ 5.99) & \textbf{117.05} ($\pm$ 6.69)\\
    	\bottomrule
    \end{tabular}

    \caption{Mean performance and standard deviation of the approaches trained on samples with up to 6 vehicles (Dataset 2).
    }
    \label{tab:gen}
\end{table*}

\subsection{Evaluation Setup}
Since high-level decision making for lane changes on highways is a very stochastic task -- the behaviour and mutual influence of the other drivers is highly unpredictable -- we evaluated our agents on a variety of different scenarios, to smooth out the high variance. We generated both three- and five-lane scenarios, varying in the number of vehicles $n \in \{30 + i \cdot 5| 0\leq i\leq 12\}$. For each fixed $n$, we evaluated 20 scenarios with different a priori randomly sampled positions and driving behaviours for each vehicle. In total, each agent was evaluated on the same 260 scenarios per fixed lane setting. The distribution of surrounding vehicles for the evaluation scenarios is shown for three and five lanes in \Cref{fig:surr_vehicles}.

\subsection{Comparative Analysis}

DeepSet-Q and Set2Set-Q were compared to a fully-connected neural network using the fixed relational grid input and a convolutional neural network using the occupancy grid. All dynamic input architectures are shown in the appendix, \Cref{tab:networks}. Additionally, we compared a naive agent with no lane changes and a rule-based controller, that uses the SUMO lane change model \textit{LC2013} with parameters shown in appendix. Each network was trained with a batch size of $64$. For optimization of all architectures, we used Adam \cite{DBLP:journals/corr/KingmaB14} with a learning rate of $10^{-4}$. The target networks were updated with $\tau=10^{-4}$. Rectified Linear Units (ReLu) were used in all hidden layers of all architectures. 
The input layers of the network architectures were optimized using Random Search with the same fixed budget for the different approaches. We preferred Random Search over Grid Search because less computational budget is needed to find models with better performance \cite{Bergstra:2012:RSH:2188385.2188395}. The final dense layers of the $Q$-module were optimized once for the fixed input architecture and kept for all other architectures. 
For the CNN architecture, we used an all-convolutional network \cite{springenberg_allnet_2015}. The configuration spaces for the Random Search are shown in the appendix, \Cref{tab:hyperopt}. 

\begin{figure*}[t!]
    \center
    \subfloat[\label{fig:more_lanes}]{%
     \includegraphics[width=0.4\textwidth,trim=0cm 0cm 0cm 2cm, clip]{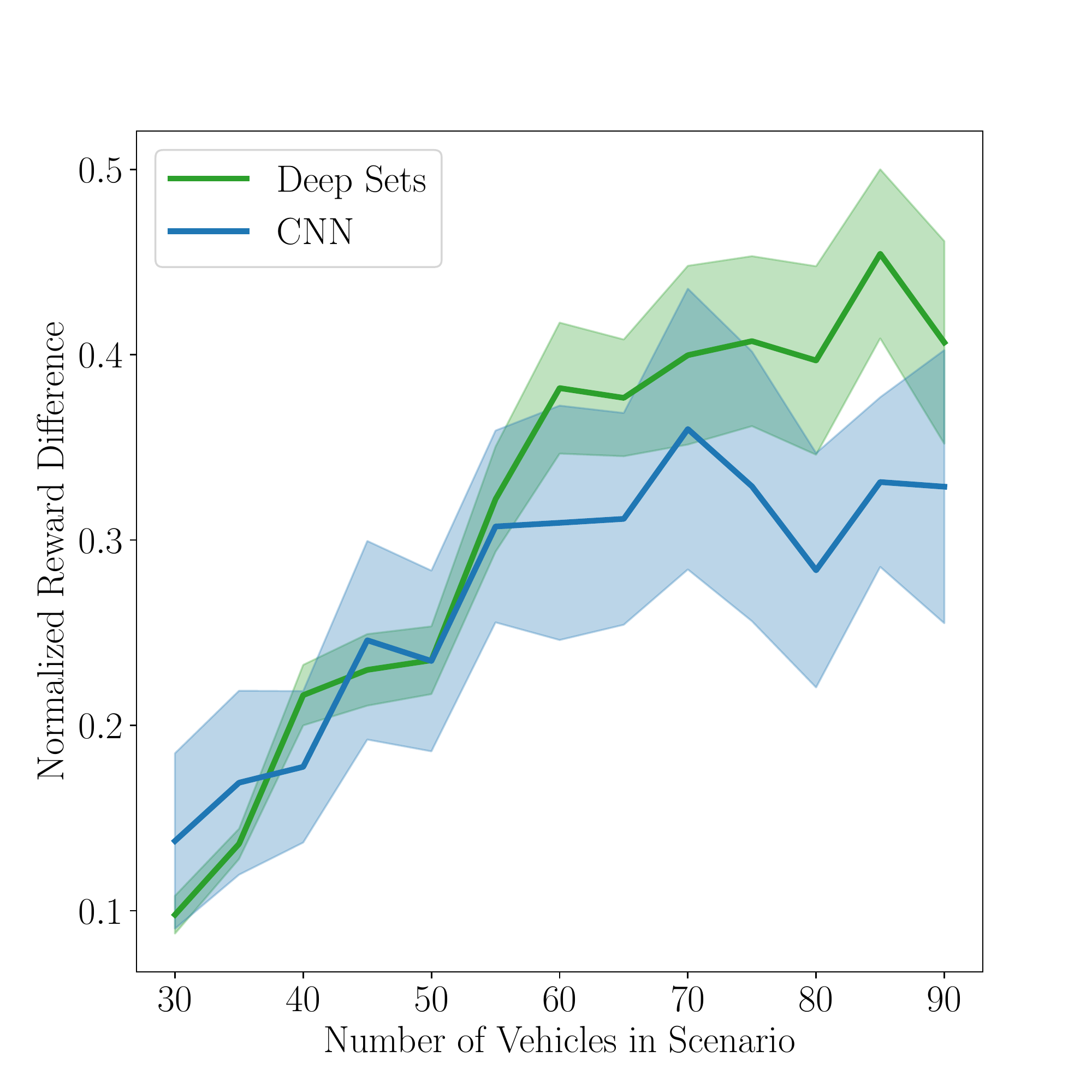}
     \hfill
    }
    \subfloat[\label{fig:noisy}]{%
    \includegraphics[width=0.4\textwidth,trim=0cm 0cm 0cm 2cm, clip]{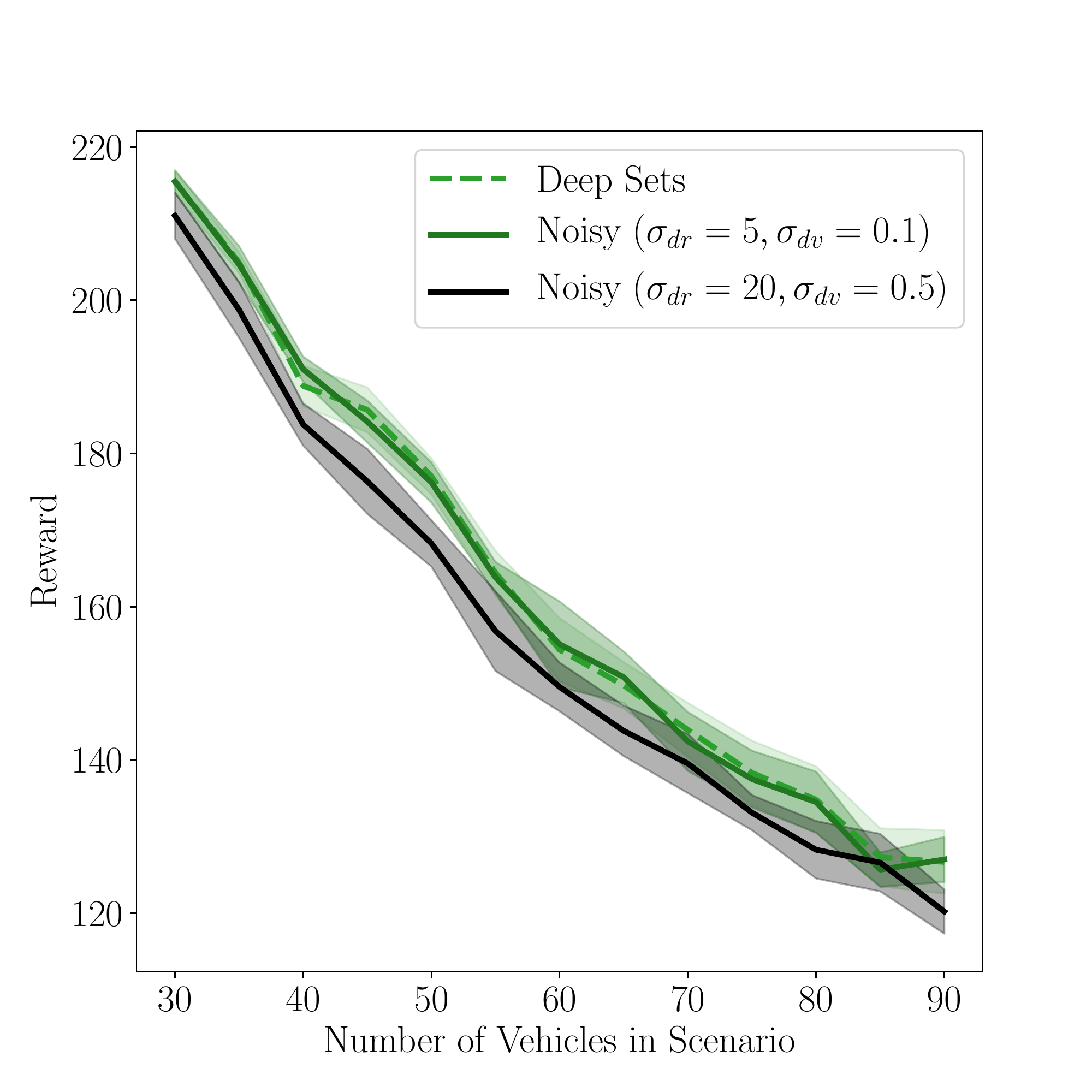}
    \hfill
   }
\caption{(a) Mean  performance and standard deviation for agents trained on scenarios with three lanes and evaluated on scenarios with five lanes. (b) Mean performance and standard deviation for DeepSet-Q agents, evaluated on noisy perception with two different noise levels.}
\end{figure*}

\section{Results}
\label{sec:results}
As can be seen in \Cref{fig:main}, the Deep Set architecture yields the best performance and lowest variance across traffic scenarios in comparison to rule-based agents and reinforcement learning agents using Set2Set, fully-connected or convolutional neural networks as input modules. For all results, see \Cref{tab:res}. A video can be found at \url{https://youtu.be/mRQgHeAGk2g}. The differences between the approaches get smaller as more vehicles are on the track due to general maneuvering limitations in dense traffic.

To investigate the generalization capabilities of all methods we trained additionally on Dataset 2, containing only transitions of at most six surrounding vehicles. As depicted in \Cref{fig:surr_vehicles}, this is only a small fraction of possible traffic situations. The evaluation results for the truncated dataset are presented in \Cref{fig:gen}. CNNs have difficulties generalizing to unseen situations and suffer from larger variance in comparison to training on Dataset 1. In contrast, Deep Sets are able to mostly keep the performance even when trained on the truncated dataset, showing only a small increase in variance. The results can be found in \Cref{tab:gen}.

Results of generalization to a differing number of lanes can be seen in \Cref{fig:more_lanes}. The plot shows the relative change in performance after increasing the number of lanes from three to five. In these situations, the CNN cannot necessarily represent the outermost lane if the agent is to the full left or right. We hypothesize this the reason for the worse performance in comparison to the Deep Sets for scenarios with a lot of vehicles on the track. In these situations also the outermost lanes can be of high importance due to the crowded traffic, but more analysis is needed to fully elucidate these differences. 

As an initial evaluation of transferability of the DeepSet-Q agent to real-world applications, we added Gaussian noise to the relative distance $dr$ and relative velocity $dv$ of all vehicles within sensor range, assuming that the relative lane can be assigned accurately. The results for two different noise levels are depicted in \Cref{fig:noisy}. Our agents show little performance decrease and seem to be robust against reasonable noise levels.

\section{Conclusion}
In this paper, we evaluated the DeepSet-Q architecture on the problem of high-level decision making in autonomous lane change maneuvers and put the results into context of current research. Deep Sets were able to outperform convolutional neural networks and recurrent attention approaches and demonstrated better generalization to unseen scenarios. Even though recent results present a connection between the bottleneck of embedding $\phi(\cdot)$ and the maximum set size \cite{2019arXiv190109006W}, we showed that Deep Sets can offer a both scalable and well-performing alternative to established approaches in the area of dynamic inputs.

Going forward, we believe there lies great potential in extending the architecture to multiple types of objects, such as pedestrians and cyclists, traffic lights or indicators for ending lanes. Further, the temporal context for traffic participants could be incorporated by including recurrent units in $\phi$. Finally, the evaluation of the architecture on a real physical system yields an important direction for future work.

\section{Acknowledgment}
We would like to thank Manuel Watter for fruitful discussions concerning dynamic input representations.

\section{Appendix}
\subsection{SUMO Configuration}
SUMO was used with a time step length of $\SI{0.5}{\second}$ and a lane change duration of $\SI{2}{\second}$. The action step length for the reinforcement learning agent is $\SI{2}{\second}$.  Acceleration and deceleration of all vehicles are  $\SI{2.6}{\metre\per\second\squared}$ and $\SI{4.5}{\metre\per\second\squared}$. The minimum gap is $\SI{2}{\metre}$, the desired time headway to $\tau=\SI{0.5}{\second}$.  As lane change controller, we use \textit{LC2013} with $\text{lcKeepRight}=0.0$. All car lengths are $\SI{4.5}{\metre}$.

\begin{table}[h]
    \centering
    \begin{tabular}{c c c c}
    \toprule
         Driver Type & maxSpeed & lcSpeedGain & lcCooperative\\
         \midrule
         Ego Driver & 24 & - & - \\
        Driver Type 1 & 24 + $u$ & $i$ & $0.0$ \\
        Driver Type 2 & 12 + $u$ & $i$ & $1.0$ \\
        Driver Type 3 & 18 + $u$ & $i$ & $0.8$ \\
        Driver Type 4 & 21 + $u$ & $i$ & $0.4$\\
        \bottomrule
    \end{tabular}
    \caption{SUMO parameters for different driver types, where $u\sim\mathcal{U}(-5, 5)$ and $i\sim\mathcal{U}(10, 20)$.}
    \label{tab:driver_types}
\end{table}

\Cref{tab:driver_types} shows the driving behaviors used to create a realistic traffic flow. Driving behaviors can be influenced by the maximum speed, the eagerness for performing lane changes to gain speed and the willingness to cooperate. For evaluation scenarios, we sampled a priori 100 different drivers uniformly from Type 1-4.\\

\subsection{Hyperparameter Optimization}
 All architectures were optimized by Random Search. The corresponding configuration spaces are shown in \Cref{tab:hyperopt}. We sampled 20 configurations, where we jointly optimized parameters and architecture layouts.
 
\begin{table}[h!]
    \centering
    \begin{tabular}{c c c}
    \toprule
    Architecture & Parameter & Configuration Space\\
    \midrule
    Fixed & Dense($\cdot$)  & $50, 100, 200$ \\
          & num layers & $2, 3$ \\
    Deep Set & $\phi$: num layers &  $1,2,3$\\
             & $\phi$: hidden/output dim & $5, 20, 100$ \\
            & $\rho$: num layers & $1,2,3$\\
             & $\rho$: hidden/output dim   & $5, 20, 100$\\
    Set2Set   & LSTM: num layers & $1, 2$\\
            & Dense($\cdot$)  &  $32, 64, 100$\\
         & iterations $T$ & $5, 20, 40$\\
  CNN & CONV: num layers & $2, 3$ \\
   &  kernel sizes& $([7, 3, 2], [2, 1])$  \\
   &  strides & $([2, 1], [2, 1])$  \\
    & filters & $8, 16, 32$  \\
       \bottomrule
    \end{tabular}   
    \caption{Configuration Spaces considered in Random Search for the Deep Set architecture and all baselines.}
    \label{tab:hyperopt}
\end{table}

\subsection{Network Architectures}

The hyperparameter-optimized architectures used in \Cref{sec:results} are shown in \Cref{tab:networks}.

\begin{table}[h!]
    \centering
    \begin{tabular}{c|c|c}
       \toprule
       CNN & Set2Set & Deep Sets \\
        \midrule
        Input($B \times 80 \times 5$)& \multicolumn{2}{c}{Input($B \times \text{seq len} \times 3$)}\\
        \midrule
         $16 \times \text{Conv2D} (3 \times 1)$ & LSTM($6$) & $\phi$: Dense($20$), Dense($80$)  \\
         $32 \times \text{Conv2D} (3 \times 1)$  & Dense($32$)&   $\rho$: Dense($80$), Dense($20$) \\
        \midrule
         \multicolumn{3}{c}{concat($\cdot$, Input($B \times 3$))}\\ 
         \multicolumn{3}{c}{Dense(100)}\\ 
         \multicolumn{3}{c}{Dense(100)}\\ 
         \multicolumn{3}{c}{Linear(3)}\\ 
         \midrule
    \end{tabular}
    \caption{Network architectures. The CNN uses strides of $(2 \times 1)$ and zero padding. For the relational grid (not shown), Input($B \times 43$) is fed directly to the Dense(100) layers.}
    \label{tab:networks}
\end{table}

\subsection{PPO}

\begin{figure}[h!]
    \centering
    \includegraphics[width=0.4\textwidth,trim=0cm 0cm 0cm 2cm, clip]{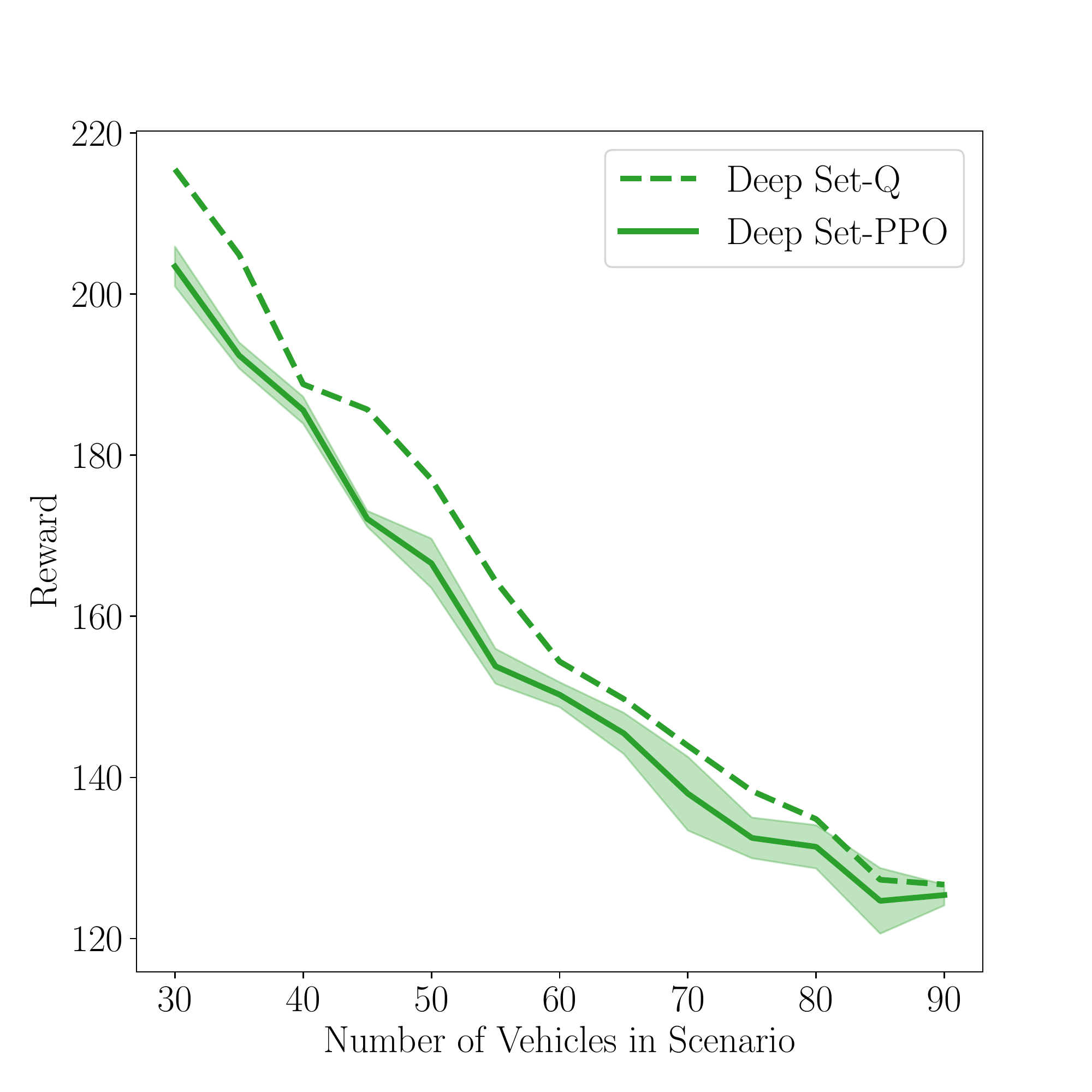}
    \caption{Mean performance and standard deviation over 7 training runs of DeepSet-PPO, compared to off-policy DeepSet-Q.}
    \label{fig:ppo}
\end{figure}

In preliminary experiments, we employed the Deep Set input in Proximal Policy Optimization (PPO) \cite{SchulmanWDRK17}. For details, see \Cref{alg:setppo}.  
However, due to the higher demand for training and the non-trivial application to autonomous driving tasks of on-policy algorithms, we switched to off-policy \mbox{Q-learning}. The performance of DeepSet-PPO is depicted in \Cref{fig:ppo}.

\begin{algorithm}[t!]
    \SetAlgoLined
    \DontPrintSemicolon
    initialize $\pi_{\mathcal{DS}}=(\phi,\rho,\pi)$ and $\pi^{old}_{\mathcal{DS}}=(\phi^{\text{old}},\rho^{\text{old}},\pi^{\text{old}})$\\
    initialize $V_{\mathcal{DS}}$\\
    
    \For{\text{optimization step} o=1,2,\dots}{
        run policy $\pi^{\text{old}}_{\mathcal{DS}}$ for $e$ episodes\\
        compute advantage estimates $\hat{A}_t$\\
        \For{\text{iteration} k=1,2,\dots, K}{
            train $V_{\mathcal{DS}}$\\
            $R_t = \frac{\pi_{\mathcal{DS}}(a_t|s_t)}{\pi^{\text{old}}_{\mathcal{DS}}(a_t|s_t)}$\\
            $R_t^{\text{CLIP}}=\text{clip}(R_t, 1-\epsilon, 1+\epsilon)$\\
            optimize surrogate: $$L^{\text{CLIP}}(\theta^{\pi_{\mathcal{DS}}})=\mathbf{\hat{E}}_t\left[\min\left(R_t\hat{A}_t, R_t^{\text{CLIP}}\hat{A}_t\right)\right]$$
        }
    }
    \caption{DeepSet-PPO}
    \label{alg:setppo}
\end{algorithm}

 The discount factor $\gamma$ was set to 0.9, since convergence to a satisfying policy needed too many updates, otherwise. We used the Clipped Surrogate Objective, with clipping threshold $\epsilon=0.2$. Value-function and policy were optimized by Adam with a learning rate of $5\cdot10^{-4}$. The Monte Carlo rollout had a horizon of 20. The batch size was set to 64.

\bibliographystyle{IEEEtran}
\clearpage
\balance
\bibliography{ref}

\begin{thebibliography}{10}
\providecommand{\url}[1]{#1}
\csname url@samestyle\endcsname
\providecommand{\newblock}{\relax}
\providecommand{\bibinfo}[2]{#2}
\providecommand{\BIBentrySTDinterwordspacing}{\spaceskip=0pt\relax}
\providecommand{\BIBentryALTinterwordstretchfactor}{4}
\providecommand{\BIBentryALTinterwordspacing}{\spaceskip=\fontdimen2\font plus
\BIBentryALTinterwordstretchfactor\fontdimen3\font minus
  \fontdimen4\font\relax}
\providecommand{\BIBforeignlanguage}[2]{{%
\expandafter\ifx\csname l@#1\endcsname\relax
\typeout{** WARNING: IEEEtran.bst: No hyphenation pattern has been}%
\typeout{** loaded for the language `#1'. Using the pattern for}%
\typeout{** the default language instead.}%
\else
\language=\csname l@#1\endcsname
\fi
#2}}
\providecommand{\BIBdecl}{\relax}
\BIBdecl

\bibitem{DBLP:journals/nature/MnihKSRVBGRFOPB15}
\BIBentryALTinterwordspacing
V.~Mnih, K.~Kavukcuoglu, D.~Silver, A.~A. Rusu, J.~Veness, M.~G. Bellemare,
  A.~Graves, M.~A. Riedmiller, A.~Fidjeland, G.~Ostrovski, S.~Petersen,
  C.~Beattie, A.~Sadik, I.~Antonoglou, H.~King, D.~Kumaran, D.~Wierstra,
  S.~Legg, and D.~Hassabis, ``Human-level control through deep reinforcement
  learning,'' \emph{Nature}, vol. 518, no. 7540, pp. 529--533, 2015. [Online].
  Available: \url{https://doi.org/10.1038/nature14236}
\BIBentrySTDinterwordspacing

\bibitem{DBLP:journals/nature/SilverHMGSDSAPL16}
\BIBentryALTinterwordspacing
D.~Silver, A.~Huang, C.~J. Maddison, A.~Guez, L.~Sifre, G.~van~den Driessche,
  J.~Schrittwieser, I.~Antonoglou, V.~Panneershelvam, M.~Lanctot, S.~Dieleman,
  D.~Grewe, J.~Nham, N.~Kalchbrenner, I.~Sutskever, T.~P. Lillicrap, M.~Leach,
  K.~Kavukcuoglu, T.~Graepel, and D.~Hassabis, ``Mastering the game of go with
  deep neural networks and tree search,'' \emph{Nature}, vol. 529, no. 7587,
  pp. 484--489, 2016. [Online]. Available:
  \url{https://doi.org/10.1038/nature16961}
\BIBentrySTDinterwordspacing

\bibitem{DBLP:conf/nips/WatterSBR15}
M.~Watter, J.~T. Springenberg, J.~Boedecker, and M.~A. Riedmiller, ``Embed to
  control: {A} locally linear latent dynamics model for control from raw
  images,'' in \emph{Advances in Neural Information Processing Systems 28:
  Annual Conference on Neural Information Processing Systems 2015, December
  7-12, 2015, Montreal, Quebec, Canada}, 2015, pp. 2746--2754.

\bibitem{DBLP:journals/jmlr/LevineFDA16}
\BIBentryALTinterwordspacing
S.~Levine, C.~Finn, T.~Darrell, and P.~Abbeel, ``End-to-end training of deep
  visuomotor policies,'' \emph{Journal of Machine Learning Research}, vol.~17,
  pp. 39:1--39:40, 2016. [Online]. Available:
  \url{http://jmlr.org/papers/v17/15-522.html}
\BIBentrySTDinterwordspacing

\bibitem{DBLP:conf/icra/JaritzCTPN18}
\BIBentryALTinterwordspacing
M.~Jaritz, R.~de~Charette, M.~Toromanoff, E.~Perot, and F.~Nashashibi,
  ``End-to-end race driving with deep reinforcement learning,'' in \emph{2018
  {IEEE} International Conference on Robotics and Automation, {ICRA} 2018,
  Brisbane, Australia, May 21-25, 2018}.\hskip 1em plus 0.5em minus 0.4em\relax
  {IEEE}, 2018, pp. 2070--2075. [Online]. Available:
  \url{https://doi.org/10.1109/ICRA.2018.8460934}
\BIBentrySTDinterwordspacing

\bibitem{adaptive_behavior_kit}
\BIBentryALTinterwordspacing
P.~Wolf, K.~Kurzer, T.~Wingert, F.~Kuhnt, and J.~M. Z{\"{o}}llner, ``Adaptive
  behavior generation for autonomous driving using deep reinforcement learning
  with compact semantic states,'' \emph{CoRR}, vol. abs/1809.03214, 2018.
  [Online]. Available: \url{http://arxiv.org/abs/1809.03214}
\BIBentrySTDinterwordspacing

\bibitem{branka_rl_highway}
B.~Mirchevska, M.~Blum, L.~Louis, J.~Boedecker, and M.~Werling, ``Reinforcement
  learning for autonomous maneuvering in highway scenarios.'' \emph{11.
  Workshop Fahrerassistenzsysteme und automatisiertes Fahren}.

\bibitem{2018TowardsPH}
M.~Nosrati, E.~A. Abolfathi, M.~Elmahgiubi, P.~Yadmellat, J.~Luo, Y.~Zhang,
  H.~Yao, H.~Zhang, and A.~Jamil, ``Towards practical hierarchical
  reinforcement learning for multi-lane autonomous driving,'' \emph{2018 NIPS
  MLITS Workshop}, 2018.

\bibitem{overtaking_maneuvers_kaushik}
M.~Kaushik, V.~Prasad, M.~Krishna, and B.~Ravindran, ``Overtaking maneuvers in
  simulated highway driving using deep reinforcement learning,'' 06 2018, pp.
  1885--1890.

\bibitem{tactical_decision_making}
M.~Mukadam, A.~Cosgun, and K.~Fujimura, ``Tactical decision making for lane
  changing with deep reinforcement learning,'' \emph{NIPS Workshop on Machine
  Learning for Intelligent Transportation Systems}, 2017.

\bibitem{deeptraffic}
L.~{Fridman}, B.~{Jenik}, and J.~{Terwilliger}, ``{DeepTraffic: Driving Fast
  through Dense Traffic with Deep Reinforcement Learning},'' \emph{arXiv
  e-prints}, p. arXiv:1801.02805, Jan. 2018.

\bibitem{2015arXiv151106391V}
O.~{Vinyals}, S.~{Bengio}, and M.~{Kudlur}, ``{Order Matters: Sequence to
  sequence for sets},'' \emph{arXiv e-prints}, p. arXiv:1511.06391, Nov. 2015.

\bibitem{Pascanu:2013:DTR:3042817.3043083}
\BIBentryALTinterwordspacing
R.~Pascanu, T.~Mikolov, and Y.~Bengio, ``On the difficulty of training
  recurrent neural networks,'' in \emph{Proceedings of the 30th International
  Conference on International Conference on Machine Learning - Volume 28}, ser.
  ICML'13.\hskip 1em plus 0.5em minus 0.4em\relax JMLR.org, 2013, pp.
  III--1310--III--1318. [Online]. Available:
  \url{http://dl.acm.org/citation.cfm?id=3042817.3043083}
\BIBentrySTDinterwordspacing

\bibitem{NIPS2017_6931}
\BIBentryALTinterwordspacing
M.~Zaheer, S.~Kottur, S.~Ravanbakhsh, B.~Poczos, R.~R. Salakhutdinov, and A.~J.
  Smola, ``Deep sets,'' in \emph{Advances in Neural Information Processing
  Systems 30}, I.~Guyon, U.~V. Luxburg, S.~Bengio, H.~Wallach, R.~Fergus,
  S.~Vishwanathan, and R.~Garnett, Eds.\hskip 1em plus 0.5em minus 0.4em\relax
  Curran Associates, Inc., 2017, pp. 3391--3401. [Online]. Available:
  \url{http://papers.nips.cc/paper/6931-deep-sets.pdf}
\BIBentrySTDinterwordspacing

\bibitem{DBLP:journals/corr/abs-1809-07480}
\BIBentryALTinterwordspacing
R.~Lee, S.~Mou, V.~Dasagi, J.~Bruce, J.~Leitner, and N.~S{\"{u}}nderhauf,
  ``Zero-shot sim-to-real transfer with modular priors,'' \emph{CoRR}, vol.
  abs/1809.07480, 2018. [Online]. Available:
  \url{http://arxiv.org/abs/1809.07480}
\BIBentrySTDinterwordspacing

\bibitem{Watkins92q-learning}
C.~J. C.~H. Watkins and P.~Dayan, ``Q-learning,'' in \emph{Machine Learning},
  1992, pp. 279--292.

\bibitem{DBLP:conf/icml/FujimotoHM18}
\BIBentryALTinterwordspacing
S.~Fujimoto, H.~van Hoof, and D.~Meger, ``Addressing function approximation
  error in actor-critic methods,'' in \emph{Proceedings of the 35th
  International Conference on Machine Learning, {ICML} 2018,
  Stockholmsm{\"{a}}ssan, Stockholm, Sweden, July 10-15, 2018}, 2018, pp.
  1582--1591. [Online]. Available:
  \url{http://proceedings.mlr.press/v80/fujimoto18a.html}
\BIBentrySTDinterwordspacing

\bibitem{Hochreiter:1997:LSM:1246443.1246450}
\BIBentryALTinterwordspacing
S.~Hochreiter and J.~Schmidhuber, ``Long short-term memory,'' \emph{Neural
  Comput.}, vol.~9, no.~8, pp. 1735--1780, Nov. 1997. [Online]. Available:
  \url{http://dx.doi.org/10.1162/neco.1997.9.8.1735}
\BIBentrySTDinterwordspacing

\bibitem{Mirchevska2018HighlevelDM}
B.~Mirchevska, C.~Pek, M.~Werling, M.~Althoff, and J.~Boedecker, ``High-level
  decision making for safe and reasonable autonomous lane changing using
  reinforcement learning,'' \emph{2018 21st International Conference on
  Intelligent Transportation Systems (ITSC)}, pp. 2156--2162, 2018.

\bibitem{SUMO}
D.~Krajzewicz, J.~Erdmann, M.~Behrisch, and L.~Bieker-Walz, ``Recent
  development and applications of sumo - simulation of urban mobility,''
  \emph{International Journal On Advances in Systems and Measurements}, vol.
  3\&4, 12 2012.

\bibitem{DBLP:journals/corr/KingmaB14}
\BIBentryALTinterwordspacing
D.~P. Kingma and J.~Ba, ``Adam: {A} method for stochastic optimization,''
  \emph{CoRR}, vol. abs/1412.6980, 2014. [Online]. Available:
  \url{http://arxiv.org/abs/1412.6980}
\BIBentrySTDinterwordspacing

\bibitem{Bergstra:2012:RSH:2188385.2188395}
\BIBentryALTinterwordspacing
J.~Bergstra and Y.~Bengio, ``Random search for hyper-parameter optimization,''
  \emph{J. Mach. Learn. Res.}, vol.~13, pp. 281--305, Feb. 2012. [Online].
  Available: \url{http://dl.acm.org/citation.cfm?id=2188385.2188395}
\BIBentrySTDinterwordspacing

\bibitem{springenberg_allnet_2015}
\BIBentryALTinterwordspacing
J.~Springenberg, A.~Dosovitskiy, T.~Brox, and M.~Riedmiller, ``Striving for
  simplicity: The all convolutional net,'' in \emph{ICLR (workshop track)},
  2015. [Online]. Available:
  \url{http://lmb.informatik.uni-freiburg.de/Publications/2015/DB15a}
\BIBentrySTDinterwordspacing

\bibitem{2019arXiv190109006W}
E.~{Wagstaff}, F.~B. {Fuchs}, M.~{Engelcke}, I.~{Posner}, and M.~{Osborne},
  ``{On the Limitations of Representing Functions on Sets},'' \emph{arXiv
  e-prints}, p. arXiv:1901.09006, Jan 2019.

\bibitem{SchulmanWDRK17}
\BIBentryALTinterwordspacing
J.~Schulman, F.~Wolski, P.~Dhariwal, A.~Radford, and O.~Klimov, ``Proximal
  policy optimization algorithms,'' \emph{CoRR}, vol. abs/1707.06347, 2017.
  [Online]. Available: \url{http://arxiv.org/abs/1707.06347}
\BIBentrySTDinterwordspacing

\end{thebibliography}

\end{document}